# What Words Do We Use to Lie?: Word Choice in Deceptive Messages

**Jason Dou**
Cornell University
CALS
xd73@cornell.edu

**Michelle Liu**
Cornell University
Engineering
mcl228@cornell.edu

**Haaris Muneer**
Cornell University
CALS
hom9@cornell.edu

**Adam Schlussel**
Cornell University
CALS
abs296@cornell.edu

### ABSTRACT

Text messaging is the most widely used form of computer-mediated communication (CMC). Previous findings have shown that linguistic factors can reliably indicate messages as deceptive. For example, users take longer and use more words to craft deceptive messages than they do truthful messages. Existing research has also examined how factors, such as student status and gender, affect rates of deception and word choice in deceptive messages. However, this research has been limited by small sample sizes and has returned contradicting findings. This paper aims to address these issues by using a dataset of text messages collected from a large and varied set of participants using an Android messaging application. The results of this paper show significant differences in word choice and frequency of deceptive messages between male and female participants, as well as between students and non-students.

### INTRODUCTION

Text messaging, or texting, has become a standard component of mobile phones and a prevalent form of conversation in society. Consequently it is now a frequent source of deceptive communication as well. We are interested in how people use words differently when they are lying, as opposed to being honest. In particular we are looking at gender and age differences between one's pronoun usage, word count, and use of noncommittal phrases. Pronouns are particularly interesting in deception because one actively chooses which pronouns he/she wants to use when communicating. Self-oriented pronouns show ownership and responsibility while other-oriented pronouns can signal distance and lack of accountability. Prior research has, among other attributes, looked at the impact of pronouns and word count on lying but returned conflicting results. Hancock et al. found that liars produced 28% more words and used fewer self-oriented pronouns (I, me) but more other-oriented pronouns (you, your) when lying than when truthful [7]. Conversely others, including Vrij [4,9,10], found that liars produce fewer words. In terms of age differences, Pennebaker & Stone found that as people age, individuals used fewer first person singular self-references, more future tense, and fewer past tense verbs [8]. This may means that people take less responsibility as they get older and focus more on the impact of events rather than what actually occurred.

### Number of Words

Prior research studied the differences between how liars and non-liars structure their statements; however, these studies have returned conflicting results. Anolli and Ciceri found liars used more words to be more persuasive and seem more credible [1]. Yet others (listed above) found that liars produce fewer words to avoid opportunities of getting caught in their own words. Burgoon's results were in the middle, finding deceivers had significantly fewer long sentences and less complex language [2].

### Pronoun Usage

Newman [8] found that across five studies, deceptive communications were characterized by fewer first-person singular pronouns, fewer third-person pronouns, more negative emotion words, fewer exclusive words, and more motion verbs. Liars used fewer first-person pronouns in an attempt to disassociate themselves from the lie [5]. Hancock also found liars used fewer self-oriented and more other-oriented pronouns when lying [6].

### Noncommittal Phrases

Noncommittal phrases are another lie indicator used so that liars do not need to commit to a certain story, leaving their intentions ambiguous. For example, liars might use "probably", "possibly", or "sure" more in deceptive messages, and message receivers can detect this [3].

***RQ1:*** *How do these message properties correlate to deception in the texts?*

> *H1.1: We expect that deceptive messages will contain more words than truthful messages.*
>
> *H1.2: We expect that liars will use fewer self-oriented and more other-oriented pronouns.*
>
> *H1.3 We expect liars to use more non-committal phrases.*

## THE PRESENT STUDY

To answer this question, we used a dataset collected via the message-sampling method by French et. al. [6], which used a custom-developed texting application that collected participants' text messages over 7 days, asking at send time whether each outgoing message was deceptive.

*Data Analysis.* In order to pare down the data set while maintaining the data's integrity, we matched participant IDs to the hashed recipient phone number to group the messages into one-sided sender conversations. We started with 1703 conversations, and after removing conversations that did not contain any lies, we were left with 351.

Of the messages remaining, the lying and truthful messages were separated. In order to evaluate the difference in amount of words between lies and truth, we averaged the word count per message for all deceptive messages and truthful messages. We did this for gender and student status as well.

For text analysis, we used a word frequency counter to count how many times each word appeared across all deceptive messages and then for all truthful messages. In order to normalize the numbers to compare between lies and truth, we divided the count number for each word over the total number of words for lies and truth respectively. We compared the percentages for self-words, other-words, and noncommittal phrases. Lastly we isolated the words that had significant differences in percentage for lies and truth.

For student status and gender, the same procedure was repeated but only for the set of text messages for each attribute. The words that were isolated for significant increase or decrease in lying messages, were compared between the attributes to see if there were any patterns or discrepancies that could possibly be explained by existing literature.

## RESULTS

### General Dataset

| Category | Avg words per text |
|---|---|
| Lie | 8.243 |
| Truth | 7. 41 |

Lying messages contain more words on average. This supports Hancock's[6] findings that liars use more words and contradicts Vrij[11], as well as supporting our hypothesis H1.1

| Word Type | In X% of deceptive messages | In X% of truthful messages |
|---|---|---|
| self-oriented | 8.3 | 11.34 |
| other-oriented | 3.87 | 4.33 |

Self words such as "i" or "i'm" were found significantly more in deceptive messages. "i" is used 0.39% more and "i'm" is used 0.34% more, which seem small, but when looking at over 36,000 words, they become more significant. Other-oriented words occur less in deceptive messages; "you" is occurs 0.58% less in deceptive messages. This finding disproves our hypothesis H1.1.

Example messages:

1. "<u>I'm</u> falling asleep!" Explanation: "just a casual lie"
2. Text: "Sorry <u>my</u> car broke down" Explanation: "I really wasn't sorry, I was annoyed because he could have walked instead of asking for a ride."

| Word Type | In X% of deceptive messages | In X% of truthful messages |
|---|---|---|
| non-committal | 1.04 | 0.59 |

The results for noncommittal phrases and qualifier statements are less definitive. "Some" and "sure" occur 0.15% more in deceptive messages. "Maybe" occurs 0.1% more and "try" occurs 0.05% more in deceptive messages. "probably" was even between honest and deceptive messages. This finding proves our hypothesis H1.3.

Example message:

1. Text: "<u>maybe</u> let me know when you are going" Explanation: "i was not going to go to the local bar to have drinks with his brother and him"

### Gender

***RQ2:*** *Do these deceptive message indicators vary according to gender?*

> *H2.1: We do not expect to see a gender difference in the word count of lies versus truths.*
>
> *H2.2: We expect females to use comparatively more other-oriented pronouns in lies than males and less self-oriented pronouns.*
>
> *H2.3: We do not expect to see a gender difference in the use of noncommittal phrases in lies versus truths.*

| Category | Avg words per text |
|---|---|
| Female - Lie | 9.177 |
| Female - Truth | 8.019 |
| Male - Lie | 7.213 |
| Male - Truth | 7.048 |

All 4 of the conditions still support Hancock et. al. [7], for both females and males: liars use more words; however, the difference for women is larger than it is for men. Women have a 12.84% increase in words per text while lying. Men overall use less words in general and only marginally (2.34%) more when lying. This partially disproves our hypothesis H2.1.

| | **Female** | **Female** | **Male** | **Male** |
|---|---|---|---|---|
| **Word type** | **In X% of deceptive messages** | **In X% of truthful messages** | **In X% of deceptive messages** | **In X% of truthful messages** |
| self-oriented | 8.91 | 7.58 | 6.43 | 5.89 |
| other-oriented | 3.56 | 4.71 | 3.69 | 4.52 |

Our data shows that frequency of "you" differs the greatest across genders. Both use the word more when telling the truth but the difference in usage is also at its greatest then. Men only use the word "you" in 4.52% while women use it 4.71% of the time in honest messages. The second greatest difference between the genders is with the word "I". Men use it 3.97% and women use it 5.31% while lying. It's interesting that when lying, female use of the word "I" increases but male use decreases. Both men and women used the words "im" and "I'm" more when lying but women used both significantly more than men.

Earlier we found that self-oriented words were used more in lying; however, when we break the data down by gender, women use self-oriented words more across the board and other-oriented words less while lying, disproving our hypothesis H2.2. Men, on the other hand, use "I" significantly less but have an increased use of "my", and "me".

However, there is a pattern with women using the deceptive indicator words significantly more than men. This could mean that men do not have as many linguistic cues for deception as women and when using linguistics to detect deception, one should weight the indicators for women more heavily than for men. One reason why men have fewer linguistic cues is that they use, on average, less words per text. For example, in the message below, the sender used the least amount of words possible to get his point across.

Example message:

1. Text from male: "def getting laid tonight" Explanation: I was trying to make our conversation longer by lying about my intentions

| | **Female** | **Female** | **Male** | **Male** |
|---|---|---|---|---|
| **Word type** | **In X% of deceptive messages** | **In X% of truthful messages** | **In X% of deceptive messages** | **In X% of truthful messages** |
| non-committal | 1.47 | 0.84 | 1.55 | 0.89 |

Not surprisingly, women and men both use non-committal phrases more when lying, proving our hypothesis H2.3. Among men,"sure" has the greatest difference and for women it is "try". Interestingly, men show a decrease in the use of "probably" while lying although we don't know why this trend exists.

Example message:

1. Female's text: "[name] gift is fine. hard to really tell. maybe it looks better in person?" Explanation: "I thought the gift(s) were so-so, but my sister was REALLY excited by them, so I didn't want to tell her that her opinion wasn't good."

### Student Status

***RQ3:*** *Do these deceptive message indicators vary by age?*

*H3.1: We do not expect to see a significant age difference in the word count of lies versus truths.*
*H3.2: We expect students to use comparatively more self-oriented pronouns in lies than non-students and less other-oriented pronouns.*

*H3.3: We do not expect to see a significant age difference in the use of noncommittal phrases in lies versus truths.*

| **Category** | **Avg words per text** |
|---|---|
| Student - Lie | 9.425 |
| Student - Truth | 7.555 |
| Non-Student - Lie | 7.642 |
| Non-Student - Truth | 7.632 |

All 4 of the conditions still support Hancock et. al. [7], for both students and non-students: people use more words when lying than when telling the truth; however, the difference for students is much more significant than for non-students. Students have a 24.75% increase in words per text when lying, while non-students only see a 0.12% increase in words. This finding disproved our hypothesis H3.1. We postulated that this extreme difference could be due to the fact that students seemed to lie about different subjects than non-students, but additional analysis will need to be performed to ascertain that other variables are not influencing the observed differences.

| | **Student** | **Student** | **Non-student** | **Non-student** |
|---|---|---|---|---|
| **Word type** | **In X% of deceptive messages** | **In X% of truthful messages** | **In X% of deceptive messages** | **In X% of truthful messages** |
| self-oriented | 8.6 | 7.56 | 7.42 | 6.6 |
| other-oriented | 2.39 | 4.26 | 4.24 | 4.29 |

Our data shows that students use significantly less other-oriented pronouns and significantly more self-oriented pronouns when lying than do non-students. This finding supported our hypothesis H3.2. Words like "I", "my", and "I'm" were found to be used more by students, but a peculiar finding was the word "me" was used slightly less by students. We are unsure why this sole self-oriented pronoun was found to be used less and will investigate it further.

| | **Student** | **Student** | **Non-student** | **Non-student** |
|---|---|---|---|---|
| **Word type** | **In X% of deceptive messages** | **In X% of truthful messages** | **In X% of deceptive messages** | **In X% of truthful messages** |
| non-committal | 1.29 | 0.612 | 0.59 | 0.50 |

Students and non-students both use non-committal phrases more when lying, although the increase is more extreme for students. Non-students saw an 18% increase in use of non-committal phrases, while students saw a massive **110.78%** increase, thus disproving our hypothesis H3.3.

There was a profound difference between students and non-students in every category of linguistic indicators of deception. Students used significantly more words, less other-oriented pronouns, and more non-committal phrases than non-students. We were a bit surprised by how large this disparity was and concluded that there may be a chance other variables influenced the trends we observed.

Overall, only four of our nine hypotheses were completely proven; however, a few of our hypotheses were chosen because of a lack of previous research related to the present study. Many of the findings from this study are novel and contribute new information to this field of research.

### Limitations and Future Work

Having access to a pre-collected dataset as opposed to performing the data collection ourselves enabled us to do more in-depth research under the given time constraints; however, there were still a few limitations we ran into. For example, due to the time limit of this study, we were only able to examine the relationship between deceptive message indicators and two of the dozens of attributes that were collected (student status and gender). In a future study, we would like to be able to look at more variables and uncover more trends. Additionally, despite this the fact that this study had a broader sample than previous studies, it was still not representative of the entire population. Certain groups were not represented equally, but the findings from this study serve as a good foundation for future work.

Additionally, there was a chance that message senders were primed to think about deception and were thus more influenced to mark a message as deceptive. Future work could include other questions after the sending of the message so that users aren't only asked about deception.

Future research could also include the use of more in-depth statistical and linguistic analysis of the dataset. In this study, word usage was examined, but there was no sentence-level linguistic analysis. Additional analysis could potentially uncover more trends and explain some of the findings that seemed unusual or unexpected. There is a s possibility that trends observed within certain groups (i.e. students using a large amount of words in lies) could be attributed to influence from other variables, additional statistical analysis should be performed in future work.